%% file: main.tex
\documentclass[10pt,twocolumn,letterpaper]{article}

\usepackage[pagenumbers]{cvpr} 


\definecolor{cvprblue}{rgb}{0.21,0.49,0.74}
\usepackage[pagebackref,breaklinks,colorlinks,allcolors=cvprblue]{hyperref}
\usepackage{listings}

\usepackage{arydshln}
\usepackage{tcolorbox} 
\usepackage{xcolor}    

\title{ANDHRA Bandersnatch: Training Neural Networks to Predict Parallel Realities}

\author{Venkata Satya Sai Ajay Daliparthi\\
Blekinge Institute of Technology\\
Karlskrona, Sweden\\
{\tt\small venkatasatyasaiajay.daliparthi@bth.se }
}

\begin{document}
\maketitle
\input{0_abstract}
\input{1_intro}

{
    \small
    \bibliographystyle{ieeenat_fullname}
    \bibliography{main}
}

\input{X_suppl}

\end{document}

%% file: 0_abstract.tex
\begin{abstract}
Inspired by Many-Worlds Interpretation (MWI), this work introduces a novel neural network architecture that splits the same input signal into parallel branches at each layer, utilizing a Hyper Rectified Activation, referred to as AND-HRA. The branched layers do not merge and form a separate network path, leading to multiple network heads for output prediction. For a network with branching factor 2 at three levels, the total heads are 2ˆ3 = 8. The individual heads are jointly trained by combining their respective loss values. However, the proposed architecture requires additional parameters and memory during training due to the additional branches. During inference, the experimental results on CIFAR-10/100 demonstrate that there exists one individual head that outperforms the baseline accuracy, achieving statistically significant improvement with equal parameters and computational cost.
\end{abstract}

%% file: 1_intro.tex
\section{Introduction}
\label{sec:intro}

As the depth of the neural networks (NN) starts increasing, the training complexity increases due to the vanishing gradient problem\cite{hochreiter1998recurrent}. As the gradients pass through each layer, they shrink, leading to an ineffective update of weights in the earlier layers (close to input). The existing solutions investigated this problem through different dimensions that include non-linear activations (ReLU \cite{nair2010rectified}), initialization techniques (Xavier \cite{glorot2010understanding} and He \cite{he2015delving}), batch normalization \cite{ioffe2015batch}, stochastic optimization (Adam \cite{KingmaB14}), and network architectures (residual \cite{he2016deep}, and dense \cite{huang2017densely} connections). In the network architectures landscape, the prominent ResNets \cite{he2016deep} introduced skip-connections between layers to facilitate direct gradient flow in deeper architectures. The DenseNet \cite{huang2017densely} connects each layer to every other layer 
thus providing each layer with direct access to gradients from all previous layers. Nevertheless, in many cases NNs are trained using a single loss function attached to the final output layer, this is due to the traditional network architecture style. To mention, some earlier works introduced methods like \textit{Companion objective} \cite{lee2015deeply}, and \textit{Auxiliary loss} \cite{szegedy2015going, larsson2016fractalnet} where an additional loss function is attached to the earlier layers for improvement in gradient flow. However, the place of these auxiliary losses remains arbitrary \cite{lee2015deeply,teerapittayanon2016branchynet}, and the auxiliary prediction is often discarded at the inference stage.

\begin{figure}
  \centering
  \begin{subfigure}{0.49\linewidth}
  \includegraphics[width=\textwidth]{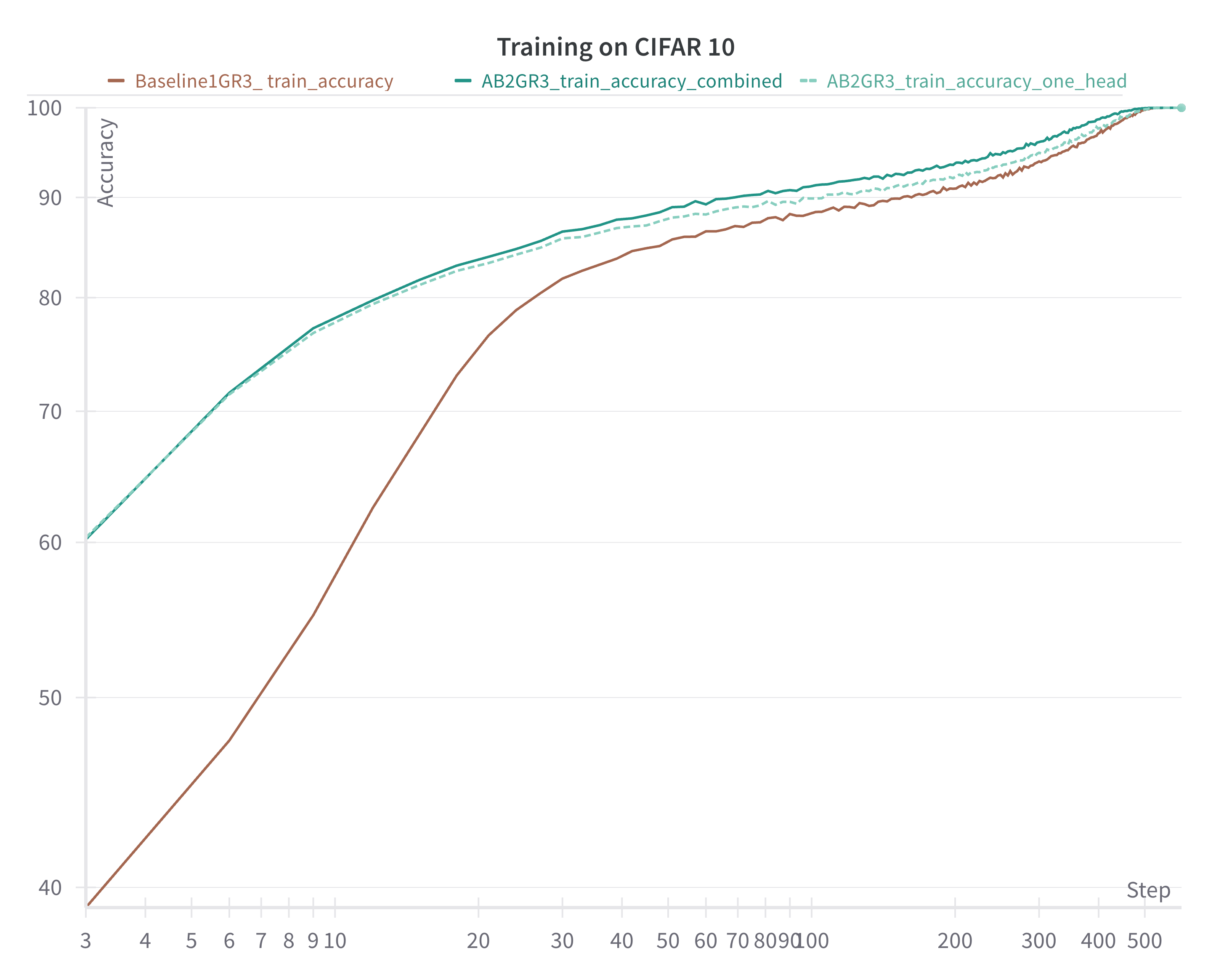}
    \label{fig:short-a}
  \end{subfigure}
  \hfill
  \begin{subfigure}{0.49\linewidth}
    \includegraphics[width=\textwidth]{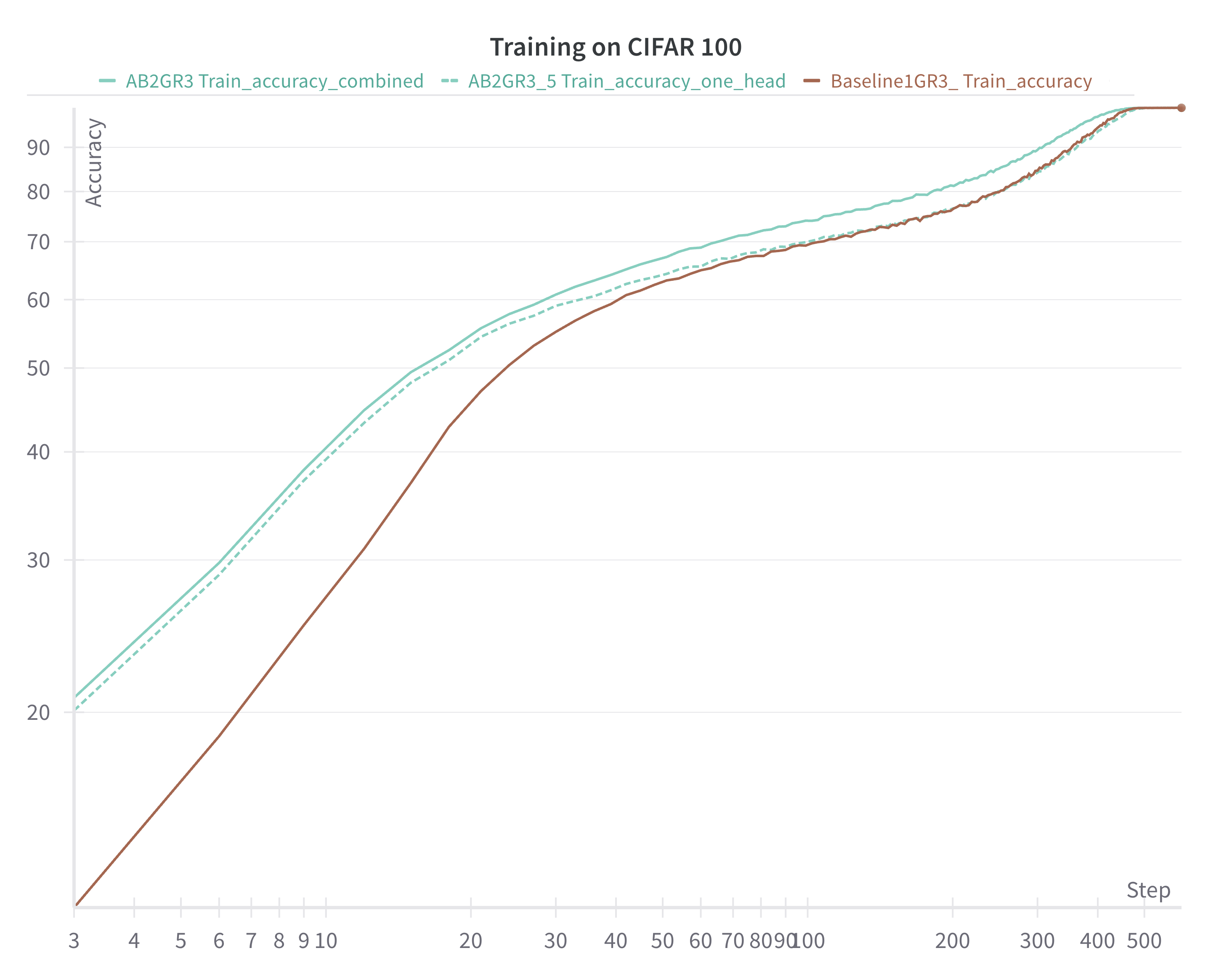}
    \label{fig:short-b}
  \end{subfigure}
  \caption{Comparison of training accuracy progression in baseline and proposed method AB (ANDHRA Bandersnatch), in log-scale graph}
  \label{fig:short}
\end{figure}

To address the vanishing gradient problem through network architectures, inspired by Many-Worlds Interpretation (MWI), this work proposes a novel NN architecture that grows exponentially by forming branches/splits at each layer where different branches independently handle the flow of information, resulting in multiple \textit{parallel} heads(output layers).
A loss function is attached to the \textit{individual heads} and the whole network is jointly trained by aggregating the individual head losses. The main contributions of this work are as follows:
\begin{itemize}
    \item A non-merging splitting/branching network module called \textit{ANDHRA}.
    \item A network architecture named \textit{ANDHRA Bandersnatch (AB)} that uses the \textit{ANDHRA} module at different levels to create network branches. 
\end{itemize}

\textit{``The key idea is that by splitting the network into multiple independent branches at each level, the flow of gradients is no longer confined to a single path. This should allow the network to effectively propagate gradients through the layers, as multiple paths are available to carry the gradient backward during training."}

The figure \ref{fig:short} presents the training accuracy progression of the proposed architecture in comparison with the baseline, where the baseline (\textit{Baseline 1GR3}) network is equivalent to a traditional feed-forward ResNet \cite{he2016deep}, and the proposed network; ANDHRA Bandersnatch (\textit{AB 2GR3}). The \textit{AB 2GR3} network has a branching factor 2 at 3 levels, the total heads for this network are  2ˆ3 = 8 heads. Here, one head in \textit{AB 2GR3} is equivalent to the baseline in terms of parameters and computational cost. Thus, in the figure \ref{fig:short},  the \textit{Baseline 1GR3} curve should be compared with \textit{AB 2GR3 one head}, and the \textit{AB 2GR3 combined} is an ensemble prediction that is inherent to the proposed architecture.

The experiential results on CIFAR-10/100 demonstrate the effectiveness of the proposed architecture by showing statistically significant accuracy improvements over the baseline networks.

\section{Method}
\label{sec:method}

This section provides a background on the source of inspiration for the proposed method, then introduces the proposed ANDHRA module, Bandersnatch network, and definition of training loss for the proposed method.

\textit{Source of Inspiration:} Many-Worlds Interpretation (MWI) of quantum mechanics assumes that every quantum measurement leads to multiple branches of reality, with each branch representing a different outcome of a quantum event. It assumes that all possible outcomes of a quantum event actually occur but in different, non-interacting branches of the universe. These parallel realities exist simultaneously, each one corresponding to a different possibility that could have occurred, leading to the idea that parallel universes are created for every quantum decision. According to MWI, a popular quantum paradox, Schrödinger Cat is interpreted as where both outcomes (the cat being dead and the cat being alive) occur, but in separate branches of the universe. There is no collapse of the wave function; the universe simply splits into two branches, one where the cat is dead and one where the cat is alive.

A similar idea of parallel realities arising from decisions (like in human choice or action, rather than purely quantum events) has been explored in various ways, often in the context of multiverse theories or alternate realities in science fiction (Netflix shows Bandersnatch and Dark).

\begin{figure}[ht!]
  \centering
  \includegraphics[width=\linewidth]{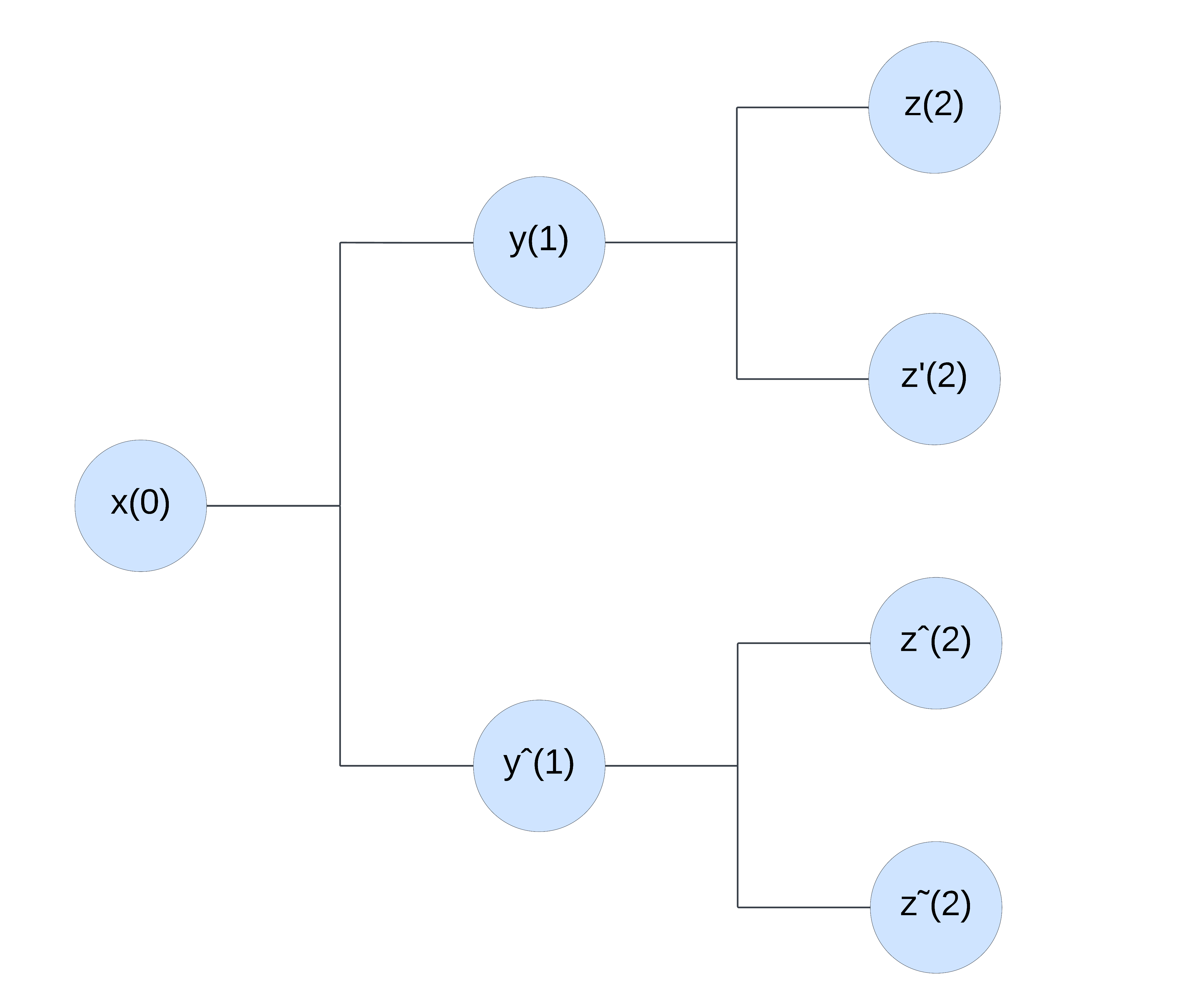}
  \caption{MWI based state changes }
  \label{fig:MWI}
\end{figure}

\subsection{Ajay N' Daliparthi Hyper Rectified Activation (ANDHRA)}

\textit{Idea: "The idea is to implement a NN architecture based on MWI where the network splits into multiple “branches” or “heads” (representing different paths) that process the same input signal in parallel, each corresponding to different possible outcomes.
Akin to how MWI suggests parallel universes in their treatment of parallelism and branching, the NN architecture involves computational paths that exist simultaneously, and those outcomes are handled independently (separate branches or worlds).", as depicted in Figure \ref{fig:MWI}}

The intuition behind the idea is that by designing a network that grows exponentially, the parent layers are shared among the individual branches, thus the shallow/earlier layers (close to input) receive multiple gradient updates from each individual branch. Since these individual branches are identical, the updates from multiple branches shouldn't deviate much from the ideal one.

\textit{Proposed method:} Based on the idea, this work proposes a network module referred to as ANDHRA that splits the given input signal into N (branching factor) number of parallel branches. The A N'D stands for Ajay and Daliparthi, and HRA stands for Hyper Rectified Activation.

Since the activation function adds non-linearity to the network, this work interprets the activation function as a decision-making point and makes a design decision to introduce the splitting function at the activation layer, the one before reducing the spatial dimensions and passing it to next-level, meaning one module for one-level.

By introducing the ANDHRA module, the network grows exponentially in terms of the number of outputs, parameters, and computational complexity. 

Let's assume that each layer uses one ANDHRA module, \( N \) is the branching factor, and \( L \) is the level of NN.

The number of heads \( H \)  at level \( L \) can be expressed as \ref{eq:Nheads} 
\begin{equation}
    H_L = N^L
  \label{eq:Nheads}
\end{equation}

The total number of layers can be expressed as the sum of the layers at each level of the network, also expressed in \ref{eq:Nheads_layer}
\begin{equation}
    \text{Layers up to level L} = H_0 + H_1 + H_2 + \ldots + H_L
  \label{eq:Nheads_layer}
\end{equation}

By substituting the formula in eq \ref{eq:Nheads} in eq \ref{eq:Nheads_layer}
\begin{equation}
    \text{Layers up to level L} = 1 + N + N^2 + N^3 + \ldots + N^L
  \label{eq:Nheads_layer_1}
\end{equation}

The equation \ref{eq:Nheads_layer_1} resembles a classic geometric series, where the first term is 1 and the common ratio is \( N \). The sum of the first \( L+1 \) terms of a geometric series is given by the formula:

\begin{equation}
    S_{L} = \frac{N^{L+1} - 1}{N - 1}
   \label{eq:geo_S}
\end{equation}

\begin{equation}
   \therefore \text{Layers up to level L } = \frac{N^{L+1} - 1}{N - 1}
  \label{eq:Nheads_layer_final}
\end{equation}

Where:
\begin{itemize}
    \item \( N \) is the branching factor.
    \item \( L \) is the Level number, starting from 0.
\end{itemize}

\subsection{ANDHRA Bandersnatch (AB) Network}

\begin{figure*}[ht!]
  \centering
  \includegraphics[width=\linewidth]{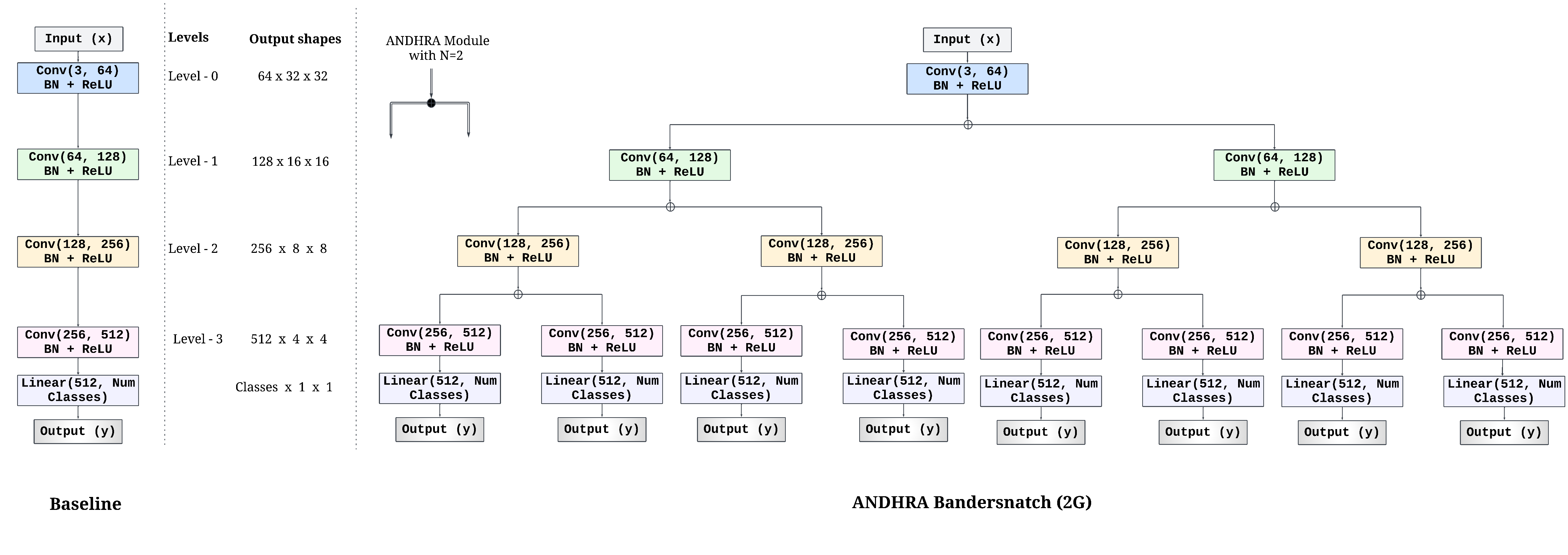}
  \caption{From the left side, baseline network, the levels \& output shapes chart, and the ANDHRA Bandersnatch 2G network}
  \label{fig:AB}
\end{figure*}

The Bandersnatch network is a NN implemented using the ANDHRA module with branching factor N = 2, denoted as ANDHRA Bandersnatch 2G (where G stands for generations also denoting network growth rate/common ratio).
It assumes that the network splits into two outcomes at each level. Based on the dataset (input image resolution), the levels will be decided in a network architecture. Figure \ref{fig:AB} presents baseline and Bandersnatch-2G network architectures side-by-side in which there are four levels (based on CIFAR input resolution 32x32), and ANDHRA module is placed three times, each at level-1, 2, and 3. The baseline architecture is implemented by replicating ResNet\cite{he2016deep}, and the Bandersnatch-2G is implemented to match the baseline for a given individual head, this can also be observed from the Figure \ref{fig:AB}. Using eq \ref{eq:Nheads}, the total heads for a 3-leveled network with branching factor 2 is 2ˆ3 = 8. Thus, the Bandersnatch-2G network consists of 8 identical heads, and the baseline is identical to an individual head in terms of parameters and computational complexity.

\textit{In Figure \ref{fig:AB}, the Conv layer at level-0 (with 3 in filters, and 64 out filters), also the first Conv layer, receives gradient updates from eight heads, the two Conv layers at level-1; each receives gradient updates from four heads, .... (the pattern repeats until the end)}

\label{ref:sec3:networknotation}
Network Notation: Each Conv block is followed by a ResBlock (R), the depth of the ResBlock will be decided during experimentation (R-Depth). A network with R0 means zero residual blocks are present in a network. For networks with R value 3, three residual blocks are stacked on top of each other, each residual block consists of two Conv layers and a skip-connection. For any given ResBlock, the number of input filters an output filters are same. 

The Conv layers represented in Figure \ref{fig:AB} have stride 2, and a point-wise (1x1 Conv) skip connection. Before passing the individual heads into linear layers, there is an average pooling layer with kernel size 4. 
Since there are 8 heads, during inference, the individual head predictions are majority-voted to get the combined prediction.

\textit{Calculating the number of layers:} using equations \ref{eq:Nheads_layer} \ref{eq:Nheads_layer_1} \ref{eq:geo_S}, the total number of layers for levels 0, 1, 2, and 3 in a Bandersnatch-2G network can be calculated as:

For each layer:
\[
H_0 = 1, \quad H_1 = 2, \quad H_2 = 4, \quad H_3 = 8
\]
The total number of Conv layers up to level 3 is:
\[
\text{Total layers up to layer 3} = 1 + 2 + 4 + 8 = 15
\]
Using the geometric sum formula:
\[
\text{Total heads up to layer 3} = \frac{2^{3+1} - 1}{2 - 1} = \frac{16 - 1}{1} = 15
\]
Thus, the total number of heads up to layer 3 is 15, this can also be manually verified by counting the number of Conv blocks at each level of the Bandersnatch-2G network in Figure \ref{fig:AB}.

\subsection{Training the ANDHRA Bandersnatch network}

While training, each head is assigned a loss function and these individual losses are combined by summing and averaging. Let \( L_1, L_2, \dots, L_N \) be the individual losses for the \( n \) heads. Each \( L_i \) corresponds to the loss computed for the \( i \)-th head of the network. The final loss \( L_{\text{total}} \) passed for back-propagation is the average of all individual losses, represented in equation \ref{eq:loss}

\begin{equation}
L_{\text{total}} = \frac{1}{n} \sum_{i=1}^{N} L_i
  \label{eq:loss}
\end{equation}

The reason for summing and averaging the losses is to create a global loss that represents the overall error across all heads. The averaging ensures that the optimization process treats each head equally, which might help avoid over-fitting to any one branch of the network, ensuring that each head contributes equally to the final loss.

For Bandersnatch Network with 8 heads, the total loss from eq \ref{eq:loss} can be written as:

\begin{equation}
L_{\text{total}} = 0.125 \cdot (L_1 + L_2 + L_3 + L_4 + L_5 + L_6 + L_7 + L_8)
  \label{eq:loss_2}
\end{equation}

\section{Evaluation}
\label{sec:eval}

\subsection{Experiment Setup}
Each network is trained five times and the mean and standard deviation values are reported.

These training hyper-parameters are kept the same for both baseline and Bandersnatch Network, and experiments are conducted by replacing just the network (The training and validation function needs adjustments to support the  Bandersnatch 2G Network): 

\begin{itemize}
    \item Dataset: CIFAR 10/100 
    \item Training data transforms: RandomCrop(32, padding=4), RandomHorizontalFlip(), and Normalize. For validation data, only Normalization.
    \item Batch Size: 128
    \item Epochs: 200
    \item Loss: CrossEntropyLoss
    \item Optimizer: SGD (momentum=0.9, weight decay=5e-4)
    \item Learning rate: 0.1
    \item Learning rate scheduler: Cosine Annealing (T max=200)
    \item Performance metric: Top-1 accuracy
\end{itemize}

\textit{Experiment Hypothesis}: Since, the baseline is identical to any individual network branch/(head) in Bandersnatch 2G Network (see Figure \ref{fig:AB}); if any individual head outperforms the baseline accuracy, during inference, that particular head can be detached and used for inference, it means improving the performance of the network without adding additional computation and parameter overhead.

To check if the experiment hypothesis holds true: a statistical significance test (Paired T-test) is performed between the results of each baseline variant and its corresponding top-performing head in Bandersnatch 2G Network. If the p-value is equal to or less than 0.05, then the prediction distributions (5 runs) are considered to be statistically significant.

\subsection{Experiment results}

In Table \ref{tab:comparison_cifar10}, and \ref{tab:comparison_cifar100}; the first column represents the depth of the residual blocks placed at each level (shown in Figure \ref{fig:AB}) of the network (refer to section \ref{ref:sec3:networknotation} network notation); the second column represents the performance of the baseline networks; the third column represents the performance of top performing heads out of the eight heads in the Bandersnatch 2G network; the fourth column represents the combined prediction of 8 heads. During the comparison, the baseline performance (col-2) is matched with the top performing head (col-3) out of 8 heads. Thus, in the fifth and sixth columns, the statistically significant difference and mean squared error is measured between the 5 runs of baseline and top performing head performance, columns (2 and 3). 

Table \ref{tab:comparison_cifar10} presents results on CIFAR-10 where the top performing head in ANDHRA Bandersnatch (2G) network outperforms the baseline from residual depth (0-3) with statistical significance difference. The \textit{experiment hypothesis} holds true in all cases, at every depth.

Table \ref{tab:comparison_cifar100} presents results on CIFAR-100 where the performance of the top performing head in ANDHRA Bandersnatch (2G) outperforms the baseline from residual depth (1-3) with a statistically significant difference. Expect, in case of residual depth (0), the proposed method slightly under-performs the baseline, thus, no statistically significant difference is observed. Hence, the \textit{experiment Hypothesis} holds true, except for row one with residual depth zero.

Furthermore, in between Table \ref{tab:comparison_cifar10} and \ref{tab:comparison_cifar100}, the performance difference is higher in Table \ref{tab:comparison_cifar100} (CIFAR-100), specifically, the rows 3 and 4 in Table \ref{tab:comparison_cifar100} with residual depths 2 \& 3, this is an interesting result, demonstrating the effectiveness of the proposed method. 
This difference can also be observed through high mean squared error in rows 3, and 4 (in Table \ref{tab:comparison_cifar100}). 

\begin{table*}[ht!]
  \centering
  \begin{tabular}{@{}lcccccc@{}}
    \toprule
    R-Depth & Baseline (1G) & \multicolumn{2}{c}{ANDHRA Bandersnatch (2G)} & Significance & Mean Sq. Error \\
    \cmidrule(lr){3-4}
    & & Top-Head & Combined & \\
    \midrule
    R0 &  93.546 $\pm$ 0.190  & \textbf{94.118 $\pm$ 0.099}  & 94.738 $\pm$ 0.090  & Yes & 0.404\\
    R1 & 95.202 $\pm$ 0.097 & \textbf{95.536 $\pm$ 0.078 } & 95.890 $\pm$ 0.099 & Yes & 0.138\\
    R2 &  95.366 $\pm$ 0.171 & \textbf{95.900 $\pm$ 0.127} &  96.230 $\pm$ 0.108  & Yes & 0.334 \\
    R3 &  95.474 $\pm$ 0.162 & \textbf{96.088 $\pm$ 0.065} &  96.378 $\pm$ 0.023  & Yes & 0.418 \\
    \bottomrule
  \end{tabular}
  \caption{Experimental results on CIFAR-10, (compare columns 2, and 3) }
  \label{tab:comparison_cifar10}
\end{table*}

\begin{table*}[ht!]
  \centering
  \begin{tabular}{@{}lcccccc@{}}
    \toprule
    R-Depth & Baseline (1G) & \multicolumn{2}{c}{ANDHRA Bandersnatch (2G)} & Significance  & Mean Sq. Error  \\
    \cmidrule(lr){3-4}
    & & Top-Head & Combined & \\
    \midrule
    R0 & \textbf{ 73.982 $\pm$ 0.184 } & 73.930 $\pm$ 0.233  & 77.186 $\pm$ 0.153  & No & 0.143\\
    R1 & 77.952 $\pm$ 0.145 & \textbf{78.792 $\pm$ 0.173 } & 81.214 $\pm$ 0.114 & Yes & 0.733\\
    R2 &  78.676 $\pm$ 0.324 & \textbf{80.354 $\pm$ 0.084} &  82.422 $\pm$ 0.113  & Yes & 2.910\\
    R3 &  78.610 $\pm$ 0.361 & \textbf{80.830 $\pm$ 0.116} &  82.784 $\pm$ 0.128  & Yes & 5.007\\
    \bottomrule
  \end{tabular}
  \caption{Experimental results on CIFAR-100, (compare columns 2, and 3)}
  \label{tab:comparison_cifar100}
\end{table*}

\section{Ablation study on ensemble prediction methods}
\label{sec:ablation}

\begin{table*}[ht!]
  \centering
  \begin{tabular}{@{}lccccc@{}}
    \toprule
    R-Depth & Majority Voting & Average Probability  & Product of Experts (PoE)&  Rank-Based Voting \\
    \midrule
    R0 &  94.738 $\pm$ 0.090   & \textbf{94.892 $\pm$ 0.110}  & 94.846 $\pm$ 0.139  & 94.818 $\pm$ 0.113\\
    
   R1 &  94.890 $\pm$ 0.099   & 96.052 $\pm$ 0.119  & \textbf{96.094 $\pm$ 0.095}  & 95.918 $\pm$ 0.098\\
   
   R2 &  96.230 $\pm$ 0.108   & \textbf{96.348 $\pm$ 0.102}  & 96.344 $\pm$ 0.096  & 96.294 $\pm$ 0.108 \\
   
  R3 &  96.378 $\pm$ 0.023   & 96.504 $\pm$ 0.108  & \textbf{96.508 $\pm$ 0.101}  & 96.428 $\pm$ 0.037\\
    \bottomrule
  \end{tabular}
  \caption{Ablation study on ensemble prediction methods of Bandersnatch network on CIFAR-10}
  \label{tab:ablation_cifar10}
\end{table*}

\begin{table*}[ht!]
  \centering
  \begin{tabular}{@{}lccccc@{}}
    \toprule
    R-Depth & Majority Voting & Average Probability  & Product of Experts (PoE)&  Rank-Based Voting \\
    \midrule
    R0 &  77.186 $\pm$ 0.153   & 77.662 $\pm$ 0.297  & \textbf{78.026 $\pm$ 0.238}  & 77.664 $\pm$ 0.218 \\

    R1 &  81.214 $\pm$ 0.114   & 81.506 $\pm$ 0.180  & \textbf{81.712 $\pm$ 0.125}  & 81.516 $\pm$ 0.132 \\

    R2 &  82.422 $\pm$ 0.113   & 82.584 $\pm$ 0.126 & \textbf{82.612 $\pm$ 0.090}   & 82.460 $\pm$ 0.119 \\

    R3 &  82.784 $\pm$ 0.128   & 82.932 $\pm$ 0.108  & \textbf{82.950 $\pm$ 0.079}  & 82.872 $\pm$ 0.138\\

    \bottomrule
  \end{tabular}
  \caption{Ablation study on ensemble prediction methods of Bandersnatch network on CIFAR-100}
  \label{tab:ablation_cifar100}
\end{table*}

Since the proposed architecture consists of multiple network predictions, the combined/ensemble prediction is used for the joint training of individual heads. Thus, an ablation study is conducted to compare different ensemble techniques on ANDHRA Bandersnatch (AB) Networks trained on CIFAR-10/100 in Section \ref{sec:eval}. Note that the default ensemble method used for the experiments in section \ref{sec:eval} is a simple majority voting.

\subsection{Selected ensemble techniques}

Let:
\begin{itemize}
    \item \( N \): Number of heads
    \item \( y_i \): Prediction of the \(i\)-th head
    \item \( p_i \): Softmax probability distribution from the \(i\)-th head
    \item \( \hat{y} \): Final combined prediction
\end{itemize}

\noindent 1. Majority Voting \cite{alex2009learning}
This strategy selects the class based on the most frequent vote among the multiple heads. By stacking all the predictions from the heads into a tensor, the mode across the predictions for each sample is calculated, as shown in Equation \ref{eq:E1} 

\begin{equation}
    \hat{y} = \text{mode}([y_1, y_2, \dots, y_N])
  \label{eq:E1}
\end{equation}

\noindent 2. Average Probability \cite{dietterich2000ensemble}

 This strategy averages the probability distributions from each head and chooses the class with the highest average probability. The probabilities from all heads are stacked, the mean is computed, and the class with the highest average probability is chosen, as shown in Equation \ref{eq:E2}

\begin{equation}
    \hat{y} = \arg\max_c \left( \frac{1}{N} \sum_{i=1}^{N} p_i[c] \right)
  \label{eq:E2}
\end{equation}

\noindent 3. Product of Experts (PoE) \cite{hinton2002training}

This strategy assumes that the heads are “experts,” and their probabilities are multiplied (in log space) to combine their opinions. The probabilities from all heads are stacked, take the log of each, sum them, and then exponentiate to get the combined probability where the class with the highest combined probability is selected, as shown in Equation \ref{eq:E32}
\begin{equation}
    \hat{y} = \arg\max_c \left( \exp \left( \sum_{i=1}^{N} \log(p_i[c] + \epsilon) \right) \right)
  \label{eq:E32}
\end{equation}

\noindent 4. Rank-Based Voting \cite{burges2005learning}

This strategy assigns higher weight to the top-ranked classes for each head. For each class, the rank scores are calculated across all heads. The ranking values are added to a tensor, where each class’s rank gets added to its corresponding position, and the class with the highest rank score is chosen. Let \( r_i[c] \) denote the rank of class \( c \) for head \( i \), the rank-based voting is shown in \ref{eq:E4}

\begin{equation}
    \hat{y} = \arg\max_c \sum_{i=1}^{N} \frac{1}{r_i[c]}
  \label{eq:E4}
\end{equation}

\subsection{Ablation study results}

From Table \ref{tab:ablation_cifar10}, the ablation results on CIFAR-10,  a similar performance is observed between the techniques; average probability and product of experts, they outperform majority voting and rank-based voting.

In Table \ref{tab:ablation_cifar100}, the ablation results on CIFAR-100, the product of experts outperforms other techniques. Similar to table \ref{tab:ablation_cifar10}, the average probability shows adequate performance.

\section{Related Work}
The Inception \cite{szegedy2015going} module proposed to split the feature map and process them with parallel convolutional layers of different kernel sizes, for capturing features at different scales. The ResNeXt\cite{xie2017aggregated} extended the ResNet \cite{he2016deep} to increase the width of the network by proposing cardinality, the number of independent splits. A similar concept of using multiple parallel convolutions has been investigated in Wide-ResNet \cite{zerhouni2017wide}, and FractalNet\cite{larsson2016fractalnet}, Res2Net \cite{gao2019res2net}. Through model architecture search methods, 
the RegNet\cite{radosavovic2020designing}, MobilenetV3 \cite{howard2019searching}, and EfficientNet \cite{tan2021efficientnetv2} balances between depth, width, and scaling.

Grouped Convolutions \cite{krizhevsky2012imagenet} is a separate branch of convolutional layers that divide the channels in an input feature map into multiple groups, and each group is processed individually, thus reducing the computational complexity of the convolutional operations. The Shufflenetv2 \cite{ma2018shufflenet}, CondenseNet \cite{huang2018condensenet}, and MobilenetV3 \cite{howard2019searching} demonstrated the effectiveness of grouped convs in designing light-weight networks. In Xception\cite{chollet2017xception}, each channel is processed independently and a 1x1 convolution is used to combine the channels, this is a special case of grouped convolution where the number of groups is equal to the channels in the input feature map.

\textit{Nevertheless, the existing works merge or concatenate feature maps after parallel processing/splitting. In contrast, this work proposes to maintain an independent branch after splitting that continues until the output layer of the network, leading to multiple network heads for prediction.}

On the other hand, the auxiliary loss \cite{szegedy2015going,teerapittayanon2016branchynet} concept proposes to introduce additional losses at intermediate layers to improve the training of earlier layers (close to input). During inference, the auxiliary heads are discarded, and the final output is considered for prediction, this can be viewed as a regularizing technique \cite{szegedy2015going}.

The concept of applying multiple loss functions is prominent in multitask learning \cite{kendall2018multi} where each loss learns to solve a specific task, these losses are combined with the primary loss for training on multiple tasks simultaneously. 

\textit{Instead, this work proposes training a network with multiple identical heads where each head is treated with a loss function and the total losses are summed and scaled before proceeding with gradient updates.}

\section{Conclusions}

This work proposes a novel NN architecture that splits the network into parallel branches where the multiple network heads are jointly trained. Due to the shared parent branches, the earlier(close to input) layers in the network receive gradient updates from multiple output heads, leading to faster convergence of the individual heads (compared to baseline as shown in Figure \ref{fig:short}). The experimental results on CIFAR-10/100 demonstrate a statistically significant difference by adopting the proposed architecture for simple ResNet style baselines. Unlike traditional methods, the ensemble prediction is inherent to the proposed architecture. Moreover, the proposed method is analogous to existing network modules, thus paving a path forward for experimentation.

%% file: X_suppl.tex
\clearpage
\maketitlesupplementary

\noindent  \textit{From the main paper results in Table \ref{tab:comparison_cifar10}, and Table \ref{tab:comparison_cifar100}, the network with residual depth three (R3) is selected for conducting additional experiments in the supplementary material. This selection is motivated by the accuracy of the networks with residual depth three. Just as in the main paper, each network is trained five times and the mean and standard deviation values are reported.}

\section{Parametric Activation}
In Figure \ref{fig:AB} (main paper), the ANDHRA module is implemented with two identical ReLU layers. However, using parametric activation functions such as PReLU, the definition of two independent layers becomes more coherent due to separate parameters for each branch. As shown in Figure \ref{fig:PReLU} where the two independent PReLU layers are defined with the number of input channels as a parameter. 

A parametric version of the baseline and the Bandersnatch -2GR3 networks are implemented by replacing the ReLU layer with PReLU (Params=input channels), and the results are presented in Table \ref{tab:PReLU}. The results demonstrate that the top performing head in Bandersnatch -2G outperforms the baseline networks in the parametric activation scenario, alining with main paper results from Table \ref{tab:comparison_cifar10}, and Table \ref{tab:comparison_cifar100}.

\begin{table*}[t!]
  \centering
  \begin{tabular}{@{}lcccccc@{}}
    \toprule
    CIFAR & Baseline (1G-PReLU) & \multicolumn{2}{c}{ANDHRA Bandersnatch (2G-PReLU)} & Significance  & Mean Sq. Error  \\
    \cmidrule(lr){3-4}
    & & Top-Head & Combined & \\
    \midrule
    10 & 95.352 $\pm$ 0.175  & \textbf{96.146 $\pm$ 0.042}  & 96.394 $\pm$ 0.069  & Yes & 0.665\\
    100 & 78.658 $\pm$ 0.504 & \textbf{80.674 $\pm$ 0.144 } & 82.584 $\pm$ 0.137 & Yes & 4.378\\
    \bottomrule
  \end{tabular}
  \caption{Parametric activation results on CIFAR10/100, }
  \label{tab:PReLU}
\end{table*}

\lstset{
    language=Python,
    basicstyle=\ttfamily\scriptsize, 
    keywordstyle=\color{blue}\bfseries,
    stringstyle=\color{red},
    commentstyle=\color{green!60!black},
    showstringspaces=false,
    breaklines=true,
    frame=single,
    numbers=left,
    numberstyle=\tiny,
    stepnumber=1,
    numbersep=5pt,
    emph={SimpleNN},
    emphstyle=\color{purple}\bfseries,
    moredelim=**[is][\color{cyan}\bfseries]{@}{@}, 
}

\begin{figure*}[!htbp] 
\begin{tcolorbox}[title= Parametric ANDHRA module, colframe=black, colback=white]
\begin{lstlisting}
class ANDHRA(nn.Module):
  def __init__(self,in_planes):
    super(ANDHRA,self).__init__()
    self.Relu1 = nn.PReLU(num_parameters=in_planes)
    self.Relu2 = nn.PReLU(num_parameters=in_planes)

  def forward(self,x):
    x1 = self.Relu1(x)
    x2 = self.Relu2(x)

    return x1, x2

\end{lstlisting}
\end{tcolorbox}
\caption{ANDHRA module with PReLU}
\label{fig:PReLU}
\end{figure*}

\begin{table*}[t!]
  \centering
  \begin{tabular}{@{}lccccc@{}}
    \toprule
    Network & \multicolumn{2}{c}{Top-1 Accuracy} & Significance  & Mean Sq. Error  \\
    \cmidrule(lr){2-3}
    & Top-Head & Combined & \\
    \midrule

    Baseline (1GR3) & 95.474 $\pm$ 0.162  & - & - & - \\

    \midrule

    AB2GR3-2H1   & 95.844 $\pm$ 0.117 & 95.670 $\pm$ 0.067 & Yes & 0.142 \\

    AB2GR3-2H2   & \textit{95.922 $\pm$ 0.150} & \textit{95.972 $\pm$ 0.104} & Yes & \textit{0.214} \\

    AB2GR3-2H3   & 95.668 $\pm$ 0.154 & 95.670 $\pm$ 0.163 & Yes & 0.084 \\

    \hdashline 

    AB2GR3-4H1   & \textbf{95.976 $\pm$ 0.151} & \textbf{96.322 $\pm$ 0.047} & Yes & \textbf{0.313} \\

    AB2GR3-4H2   & 95.906 $\pm$ 0.160 & 95.882 $\pm$ 0.170 & Yes & 0.249 \\

   \bottomrule
  \end{tabular}
  \caption{Ablation study results on CIFAR-10 for ANDHRA module at different levels}
  \label{tab:ablation_module_10}
\end{table*}

\begin{table*}[t!]
  \centering
  \begin{tabular}{@{}lccccc@{}}
    \toprule
     Network & \multicolumn{2}{c}{Top-1 Accuracy} & Significance  & Mean Sq. Error  \\
    \cmidrule(lr){2-3}
    & Top-Head & Combined & \\
    \midrule
    Baseline (1GR3) &  78.610 $\pm$ 0.361 & - & - & - \\
    \midrule

    AB2GR3-2H1   & 79.660 $\pm$ 0.260 & 79.674 $\pm$ 0.182 & Yes & 1.130 \\

    AB2GR3-2H2   & \textit{80.100 $\pm$ 0.200} & \textit{ 80.140 $\pm$ 0.036} & Yes & \textit{2.301} \\

    AB2GR3-2H3   & 79.444 $\pm$ 0.143 & 79.380 $\pm$ 0.116 & Yes & 0.747 \\

    \hdashline 
    AB2GR3-4H1   & \textbf{80.484 $\pm$ 0.141} & \textbf{82.188 $\pm$ 0.260} & Yes & \textbf{3.621} \\

    AB2GR3-4H2   & 80.294 $\pm$ 0.087 & 81.324 $\pm$ 0.299 & Yes & 2.991 \\

    \bottomrule
  \end{tabular}
  \caption{Ablation study results on CIFAR-100 for ANDHRA module at different levels}
  \label{tab:ablation_module_100}
\end{table*}

\section{ANDHRA module at different levels}
In the main paper, for the network Bandersnatch 2G (refer Figure \ref{fig:AB}, one ANDHRA module is placed at each network level starting from level 1-3. Thus, the network in Figure \ref{fig:AB} consists of three ANDHRA modules, leading to 8 output heads. In this section, an ablation study is performed with:

\begin{enumerate}
    \item One ANDHRA module = 2 network heads
    \item Two ANDHRA modules = 4 network heads
\end{enumerate}

\subsection{One ANDHRA module and 2 output heads}

\begin{figure*}[ht!]
  \centering
  \includegraphics[width=\linewidth]{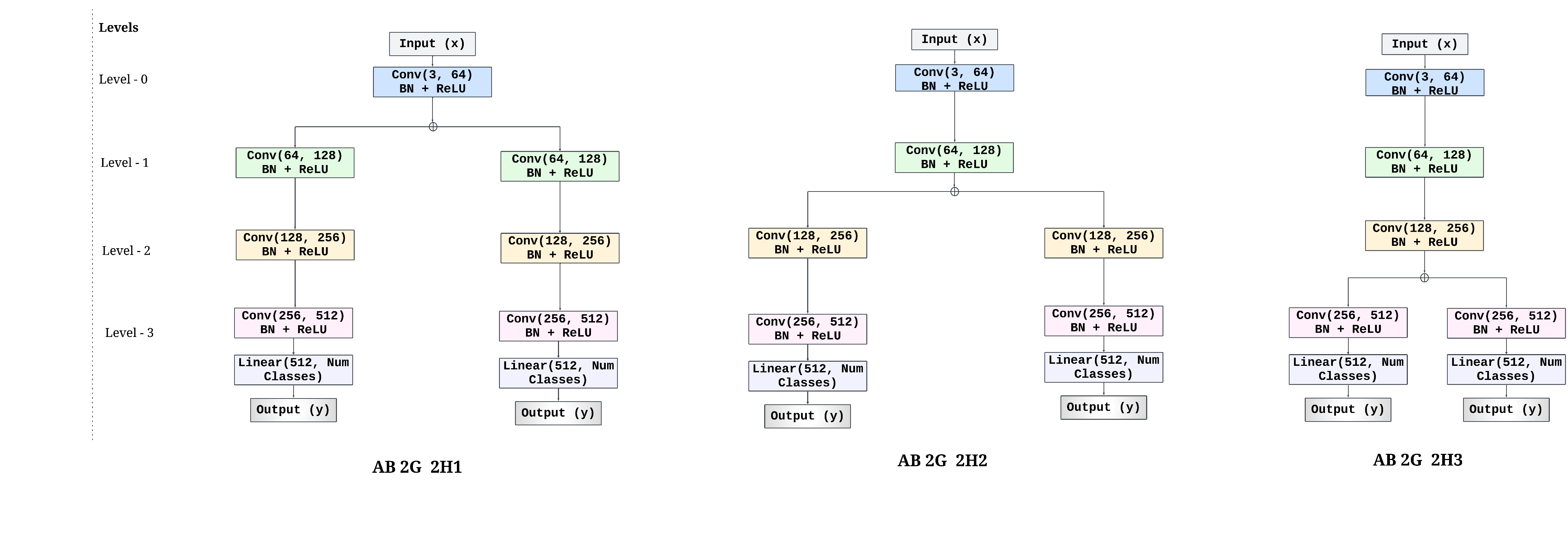}
  \caption{From the left side: levels chart, AB2GR3-2H1, AB2GR3-2H2, and AB2GR3-2H3 networks}
  \label{fig:AB2H}
\end{figure*}

Since there are three possibilities of placing the ANDHRA module at levels (1, 2, and 3), \textit{three networks (AB2GR3-2H1, AB2GR3-2H2, and AB2GR3-2H3)} are implemented as shown in the Figure \ref{fig:AB2H}.

Note: the network code presented in Figure \ref{fig:network2H} belongs to this family of networks with one ANDHRA module placed at level 1. (AB2GR\textbf{1}-2H1)

\subsection{Two ANDHRA modules and 4 output heads}

\begin{figure*}[ht!]
  \centering
  \includegraphics[width=\linewidth]{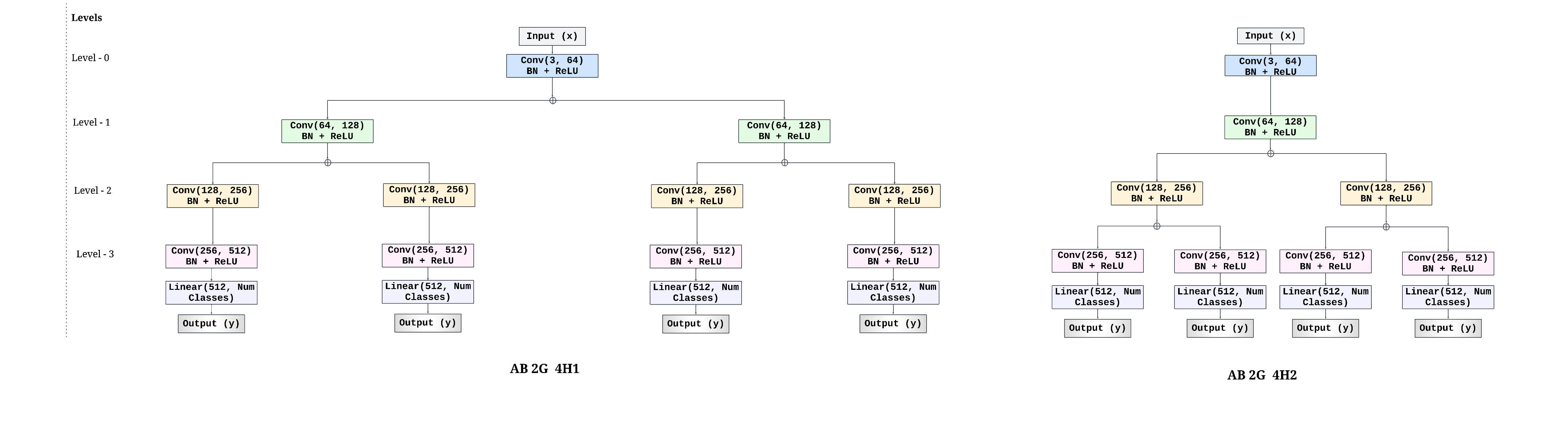}
  \caption{From the left side: levels chart,  AB2GR3-4H1, and AB2GR3-4H2 networks}
  \label{fig:AB4H}
\end{figure*}

Since there are two possibilities of placing 2 ANDHRA modules at levels (1-2, and 2-3), \textit{two networks (AB2GR3-4H1 and AB2GR3-4H2)} are implemented as shown in the Figure \ref{fig:AB4H}.

\subsection{Results}

The total 5 five networks (3 two heads - 2H) + 2 four heads - 4H) are trained on CIFAR-10/100, and the results are presented in Table \ref{tab:ablation_module_10}, and Table \ref{tab:ablation_module_100} along with the baseline network (from main paper, baseline with ReLU). The statistical significance test is performed between the baseline and top-performing head in the Bandersnatch network.

In Table \ref{tab:ablation_module_10} and Table \ref{tab:ablation_module_100}, all the Bandersnatch 2G variants (2H, 4H) outperformed the baseline network in terms of top-1 accuracy with statistically significant difference. Further, the network AB2GR3-4H1 outperforms out of the five Bandersnatch network variants trained in this ablation study.

\section{Implementation}
\label{sec:Implementation}

This section presents the implementation of the Bandersnacth-2G Network through a minimal network with the ANDHRA module placed only at level 1, meaning splitting is performed only once, thus leading to 2 output heads. In this network, the residual module depth is limited to one (R1). The PyTorch code for implementing this minimal network is presented in three parts (in  figures \ref{fig:modules}, \ref{fig:network2H}, and \ref{fig:TrainingLoop}):
\begin{enumerate}
    \item \textit{Network Modules (\ref{fig:modules})}: consists of three building blocks of the network that include the ANDHRA module, a residual module with depth-1, and a residual module for pooling and feature space expansion.
    \item \textit{Bandersnatch 2G network with 2 heads (\ref{fig:network2H})}: consists of network definition and forward-pass where the ANDHRA module is only placed at level-1, and the network returns two outputs.

    \item  \textit{Training function (\ref{fig:TrainingLoop})} consists of combined loss and majority voting prediction out of two output heads.
    
 \end{enumerate}

\begin{figure*}[!htbp] 
\begin{tcolorbox}[title=Network Modules, colframe=black, colback=white]
\begin{lstlisting}
class @ANDHRA@(nn.Module):   # Proposed splitting module 
  def __init__(self):
    super(ANDHRA,self).__init__()
    self.Relu1 = nn.ReLU(inplace=False)
    self.Relu2 = nn.ReLU(inplace=False)

  def forward(self,x):
    x1 = self.Relu1(x)
    x2 = self.Relu2(x)

    return x1, x2

class ResBlock(nn.Module): # Residual block with equal in/out filters
    def __init__(self, in_planes):
        super(ResBlock3, self).__init__()

        #residual function
        self.conv = nn.Sequential(
            nn.Conv2d(in_planes, in_planes, kernel_size=3, stride =1,padding=1, bias=False),
            nn.BatchNorm2d(in_planes),
            nn.ReLU(inplace=False),
            nn.Conv2d(in_planes,in_planes, kernel_size=3, stride =1,padding=1, bias=False),
            nn.BatchNorm2d(in_planes))
            
        #shortcut
        self.shortcut = nn.Sequential()

    def forward(self, x):
        out = self.conv(x)
        out += self.shortcut(x)
        return out

class ResBlockP(nn.Module): # Residual block with inherent pooling that also doubles in filters
    def __init__(self, in_channels, out_channels, stride):
        super(ResBlockP, self).__init__()

        #residual function
        self.residual_function = nn.Sequential(
            nn.Conv2d(in_channels, out_channels, kernel_size=3, stride= stride, padding=1, bias=False),
            nn.BatchNorm2d(out_channels),
            nn.ReLU(inplace=False),
            nn.Conv2d(out_channels, out_channels, kernel_size=3, padding=1, bias=False),
            nn.BatchNorm2d(out_channels)
        )

        #shortcut
        self.shortcut = nn.Sequential(
              nn.Conv2d(in_channels, out_channels, kernel_size=1, stride=stride, bias=False),
              nn.BatchNorm2d(out_channels)
        )

    def forward(self, x):
        return nn.ReLU(inplace=False)(self.residual_function(x) + self.shortcut(x))

\end{lstlisting}
\end{tcolorbox}
\caption{Modules of the network}
\label{fig:modules}
\end{figure*}

\begin{figure*}[!htbp] 
\begin{tcolorbox}[title= Bandersnatch 2G network with 2 heads, colframe=black, colback=white]
\begin{lstlisting}
class AB_2GR1_2H1(nn.Module): 
  def __init__(self, num_classes):
    super(AB_2GR1_2H1,self).__init__()
    self.Conv1 =  nn.Sequential(
                    nn.Conv2d(3, 64, kernel_size=3, stride =1,padding=1, bias=False), 
                    nn.BatchNorm2d(64),
                    nn.ReLU(inplace=False))
    self.Res1 = ResBlock(in_planes = 64)

    @self.Act1 = ANDHRA()@ # Proposed splitting module

    self.Conv21  = ResBlockP(in_channels=64, out_channels=128, stride=2) # Branch 1
    self.Res21   = ResBlock(in_planes = 128)

    self.Conv22  = ResBlockP(in_channels=64, out_channels=128, stride=2) # Branch 2
    self.Res22   = ResBlock(in_planes = 128)
    
    self.Act21   = nn.ReLU(inplace=False)
    self.Act22   = nn.ReLU(inplace=False)

    self.Conv31  = ResBlockP(in_channels=128, out_channels=256, stride=2)
    self.Res31   = ResBlock(in_planes = 256)

    self.Conv32  = ResBlockP(in_channels=128, out_channels=256, stride=2)
    self.Res32   = ResBlock(in_planes = 256)                            
    
    self.Act31   = nn.ReLU(inplace=False)
    self.Act32   = nn.ReLU(inplace=False)

    self.Conv41  = ResBlockP(in_channels=256, out_channels=512, stride=2)
    self.Res41   = ResBlock(in_planes = 512)
    
    self.Conv42  = ResBlockP(in_channels=256, out_channels=512, stride=2)
    self.Res42   = ResBlock(in_planes = 512)

    self.Relu    = nn.ReLU(inplace=False)
    self.pool4   = nn.AvgPool2d(kernel_size=4) 
    
    self.Linear1 = nn.Linear(512, num_classes)
    self.Linear2 = nn.Linear(512, num_classes)

  def forward(self,x):
  
    out = self.Res1(self.Conv1(x))

    @out1, out2 = self.Act1(out)@ # Splitting at level 1

    out1 = self.Res21(self.Conv21(out1)) # Branch 1
    out2 = self.Res22(self.Conv22(out2)) # Branch 2
    
    out1 = self.Act21(out1)
    out2 = self.Act22(out2)

    out1 = self.Res31(self.Conv31(out1))
    out2 = self.Res32(self.Conv32(out2))
    
    out1 = self.Act31(out1)
    out2 = self.Act32(out2)

    out1 = self.Linear1(self.pool4(self.Relu(self.Res41(self.Conv41(out1)))).view(out.size(0), -1))
    out2 = self.Linear2(self.pool4(self.Relu(self.Res42(self.Conv42(out2)))).view(out.size(0), -1))

    return out1, out2 

\end{lstlisting}
\end{tcolorbox}
\caption{Network initialization and forward-pass, ANDHRA module is only placed at level 1}
\label{fig:network2H}
\end{figure*}

\begin{figure*}[!htbp] 
\begin{tcolorbox}[title=Training function with Combined Loss and Majority Voting, colframe=black, colback=white]
\begin{lstlisting}
def train(epoch): # Training function
    print('\nEpoch: %d' % epoch)
    net.train()
    train_loss = 0
    correct = 0
    total = 0
    
    # Initialize counters for individual model accuracies
    correct_individual = [0] * 2
    total_individual = 0

    for batch_idx, (inputs, targets) in enumerate(trainloader):
        inputs, targets = inputs.to(device), targets.to(device)
        optimizer.zero_grad()
        @out1, out2 = net(inputs)@

        # Calculate losses for each output
        @loss1 = criterion(out1, targets)@
        @loss2 = criterion(out2, targets)@
        
        # Combine losses and backpropagate
        @loss =  0.5 * (loss1 + loss2)@
        loss.backward()
        optimizer.step()

        train_loss += loss.item()
        
        # Predictions and majority voting
        outputs = [out1, out2]
        individual_predictions = [output.max(1)[1] for output in outputs]
        
        # Majority vote prediction
        p = torch.stack(individual_predictions, dim=0).cpu().detach().numpy()
        @m = stats.mode(p)@
        predicted_majority = torch.from_numpy(m[0]).squeeze().cuda()
        
        # Update majority correct count
        total += targets.size(0)
        correct += predicted_majority.eq(targets).sum().item()

        # Update individual model correct counts
        for i, pred in enumerate(individual_predictions):
            correct_individual[i] += pred.eq(targets).sum().item()
        total_individual += targets.size(0)

\end{lstlisting}
\end{tcolorbox}
\caption{Training function with Combined Loss and Majority Voting}
\label{fig:TrainingLoop}
\end{figure*}